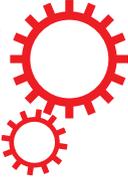

# SCIENTIFIC REPORTS

# Interactive Volumetry Of Liver Ablation Zones

Jan Egger[1,2,3], Harald Busse[4], Philipp Brandmaier[4], Daniel Seider[4], Matthias Gawlitza[4], Steffen Strocka[4], Philip Voglreiter[2], Mark Dokter[2], Michael Hofmann[2], Bernhard Kainz[5], Alexander Hann[6], Xiaojun Chen[7], Tuomas Alhonnoro[1], Mika Pollari[1], Dieter Schmalstieg[2,*] & Michael Moche[4,*]

Percutaneous radiofrequency ablation (RFA) is a minimally invasive technique that destroys cancer cells by heat. The heat results from focusing energy in the radiofrequency spectrum through a needle. Amongst others, this can enable the treatment of patients who are not eligible for an open surgery. However, the possibility of recurrent liver cancer due to incomplete ablation of the tumor makes post-interventional monitoring via regular follow-up scans mandatory. These scans have to be carefully inspected for any conspicuousness. Within this study, the RF ablation zones from twelve post-interventional CT acquisitions have been segmented semi-automatically to support the visual inspection. An interactive, graph-based contouring approach, which prefers spherically shaped regions, has been applied. For the quantitative and qualitative analysis of the algorithm's results, manual slice-by-slice segmentations produced by clinical experts have been used as the gold standard (which have also been compared among each other). As evaluation metric for the statistical validation, the Dice Similarity Coefficient (DSC) has been calculated. The results show that the proposed tool provides lesion segmentation with sufficient accuracy much faster than manual segmentation. The visual feedback and interactivity make the proposed tool well suitable for the clinical workflow.

Liver cancer is on the rise worldwide, mainly because of hepatitis infection and alcohol abuse. Especially patients with primary liver cancer (Hepatocellular Carcinomas, HCC) have a poor prognosis because, of its late symptomatic onset, resulting in a median survival time of four to six months from the time of diagnosis, when untreated. According to the recent treatment guidelines[1], radiofrequency ablation (RFA) serves as a first line therapy approach for early HCC in patients with liver cirrhosis. Also, for metastatic liver disease, the local usage of ablation therapies increases. While the technique was originally developed for patients who were not eligible for surgery its use has now expanded to serve as a bridge to liver transplantation and even as an alternative to surgical resection in the early stages of the disease[2]. RFA was first described in the early 1990s, followed by huge technical advances throughout the last decades. The underlying principle is based on a high frequency alternating current, which is delivered through one or more electrodes in the treated lesion[3] (see Fig. 1 for a schematic view of RFA needle placement in a liver tumor, including the surrounding necrotic zone, and Fig. 2 for a postinterventional computed tomography (CT) scan with ablation needle). Optimally, the heat destroys cancer cells by inducing a coagulative necrosis, with cellular proteins being denaturized. Most commonly, tissue necrosis already begins at approximately 60 °C, but usually, temperatures around 100° are needed to achieve satisfying results. The amount of destroyed tissue mostly depends on the individual impedance and placement of the needle. Furthermore, it is inversely proportional to the square of the distance from the electrode. As a result, tissue cools rapidly away from the tip of the needle probe. Hence, the proximity to large blood vessels also plays a major role in the heat transmission. Blood flow protects the vessel wall from damage, but, on the flipside, acts as a heat sink by cooling down nearby tissue limiting the methods overall success[4]. As a consequence, a significant mismatch between expected and truly induced lesion size and geometry has been observed in many radio-frequency ablations performed in the liver. It can lead to over-treatment with severe injuries (up to 9% major complications[5]), or under-treatment with tumor recurrence (up to 40%[5]).

[3]Department of Neuroscience and Biomedical Engineering, Aalto University, Rakentajanaukio 2 C, 02150 Espoo, Finland. [2]Institute for Computer Graphics and Vision, Faculty of Computer Science and Biomedical Engineering, Graz University of Technology, Inffeldgasse 16, 8010 Graz, Austria. [3]BioTechMed-Graz, Austria. [4]Department of Diagnostic and Interventional Radiology, Leipzig University Hospital, Liebigstraße 20, 04103 Leipzig, Germany. [5]Department of Computing, Imperial College London, Huxley Building, 180 Queen's Gate, London SW7 2AZ, UK. [6]Department of Internal Medicine and Gastroenterology, Katharinenhospital, Kriegsbergstraße 60, 70174 Stuttgart, Germany. [7]Institute of Biomedical Manufacturing and Life Quality Engineering, School of Mechanical Engineering, Shanghai Jiao Tong University, Dong Chuan Road 800, Shanghai Post Code: 200240, China. *These authors jointly supervised this work. Correspondence and requests for materials should be addressed to J.E. (email: egger@tugraz.at)

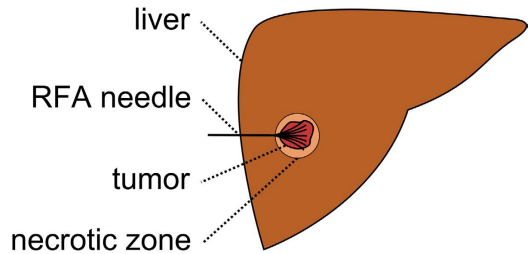

**Figure 1. Schematic view of the liver (brown) with a fully expanded and umbrella-shaped radiofrequency ablation (RFA) needle (black).** The needle tips are located in a liver tumor (red) surrounded by the so called necrotic zone (light brown).

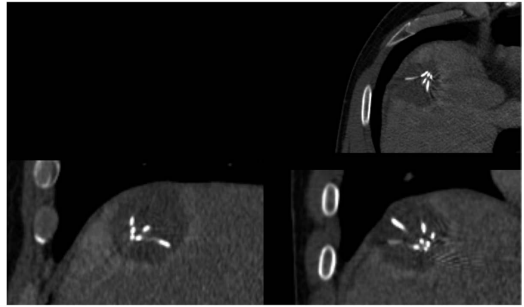

**Figure 2. Postinterventional computed tomography (CT) scan of a radiofrequency (RF) ablation with the ablation needle still in place: the upper right window shows an axial plane, the lower left window a sagittal plane and the lower right window a coronal plane.** The RFA needle is easily recognizable, because its umbrella-shaped characteristics show up very bright within the ablation zone inside the liver.

Tumor recurrence has a major limitation on the survival rates for all types of cancer therapies, i.e., resection and RFA. Cohort patient studies have shown the evidence of a significant reduction in the recurrence rate, if the RFA generates a safety rim around the tumor[6]. This elicits the need for a reliable method for the comparison of localization, size and geometry of the tumor in the preoperative images, on the one hand, and thermally induced lesion after ablation, on the other hand.

Tumor recurrence can be diagnosed by the detection of typical alteration in tissue enhancement. Nevertheless, increase of size or geometry of the lesion seems to be a much more sensitive indicator of early recurrence in follow up imaging. Therefore, a reliable and feasible determination of the ablation zone at baseline and follow-up may contribute to a positive outcome for the patient and can lead to a better understanding about the cause of new tumor growth. Consequently, the additional knowledge might lead to improvement of ablation protocols or even new treatment strategies.

Determination of therapeutically induced lesions after minimally invasive cancer treatment can be performed by segmentation. This is usually done by a time consuming manual procedure and, therefore, not yet part of the clinical routine. A validated segmentation algorithm may potentially increase the acceptance of the method in the medical community and, consecutively, lead to a significant benefit in patient treatment.

The *segmentation* field in computer vision deals with the computer-aided analysis and classification of (medical) image data in a broad range of applications, such as the automatic detection of humans in videos[7] or the automatic volumetry of brain images of patients[8]. In general, the objective of a segmentation algorithm is to support and speed-up a time-consuming manual selection and contouring process.

A number of algorithms have been proposed, i.e., Active Contours (ACM)[9,10], Deformable Models[11,12] Active Appearance Models (AAM)[13,14], graph-based approaches[15,16], fuzzy-based approaches[17], or neural networks[18], which are often based on a mathematical model from other disciplines, like Physics or Electrical Engineering, or present a combination of several mathematical models. Alternatively, the algorithms can be classified either as fully-automatic or semi-automatic. In the latter, the user is able to guide the segmentation algorithm to avoid an unsatisfying segmentation outcome. In contrast, fully-automatic approaches generally need a re-run after failure (e.g., with other parameter settings), which can be a very frustrating trial-and-error process for the end-user. An example of a fully-automatic liver tumor segmentation approach for abdominal computed tomography scans has been introduced by Abdel-massieh et al.[19]. The approach does not require any interaction from the user, but applies Gaussian smoothing and isodata thresholding to turn the gray value image into a binary representation, with tumors visible as black spots on a white background. Another fully-automatic liver tumor segmentation algorithm using Fuzzy Generalized C-Means (FGCM) and Possibilistic Generalized C-Means (PGCM) has been presented by Mandava et al.[20]. This kernel based clustering algorithm incorporates Tsallis entropy to resolve long-range interactions between tumor and healthy tissue intensities and uses the datasets from the MICCAI liver Tumor Segmentation Challenge 08 (LTS08) for evaluation.

As we propose a semi-automatic approach in the area of liver tumor treatment, we will introduce more background work within this field. A 3D fuzzy-based approach to liver tumor segmentation has been introduced by Badura and Pietka[21]. Their semi-automatic liver segmentation scheme consists of a seed point selection, three-dimensional anisotropic diffusion filtering, and adaptive region growing, supported by a fuzzy inference system. Zhang et al.[22] have proposed an interactive tumor segmentation method for CT scans from the liver using Support Vector Classification (SVM) with Watershed. In summary, they partition the CT volume under a watershed transform and some pre-processing operations, followed by an SVM classifier trained on the user-selected seed points. Moreover, morphological operations are performed to refine the rough segmentation result of the SVM classification. A related SVM-based semi-automatic method for liver tumors from computed tomography angiography (CTA) images using voxel classification and affinity constraint propagation has been presented by Freiman et al.[23]. Their method employs user-defined seeds to classify the liver voxels into tumor and healthy tissue groups, supported by an SVM classification engine. In a following step, an energy function describing the propagation of these seeds is defined over the 3D image. Another semi-automatic segmentation approach is to use Bayesian rule-based 3D region growing, as has been proposed by Qi et al.[24]. They initialize a bag of Gaussians at manually selected seeds, which they iteratively update during the growing process. Additionally, morphological operations are performed to refine the result afterwards, and, finally, to obtain a binary segmentation, the related continuous segmentation map is thresholded. Closer related to our work is the work of Häme and Pollari[25], who have presented a liver tumor segmentation method to reduce the manual labor and time required in the treatment planning of radiofrequency ablation. To achieve this, they introduced a semi-automatic liver tumor segmentation approach with a hidden Markov measure field model and a non-parametric distribution estimation.

Others working in the specific area of (semi-)automatic segmentation of ablation zones in RFA datasets of the liver are Passera et al.[26]. They claim to present the first attempt to obtain a quantitative tool aimed to assess the accuracy of RFA. For the segmentation, they use a Live-Wire algorithm[27,28] – implemented within the MeVisLab platform (www.mevislab.de) – and clustering. However, they did not include RFA data in their study if the needle was still present within the scan. But this is the case when the interventionalist particularly wants to assess the size of the induced lesion under the assumption of continuation or additional ablation. And to avoid the repositioning or the replacement of the instrument, the RFA probe remains in the patient while performing the control scan. Additionally, the segmentation approach worked only in 2D, which can be very time-consuming in case of tumors or ablation zones that extend over many slices; the reported segmentation time was ten minutes. A separate radio frequency ablation registration, segmentation, and fusion tool called RFAST has been presented by McCreedy et al.[29], who also apply a livewire-based method in single 2D slices. However, the segmentation process has not been described in detail and no quantitative segmentation results are presented. Keil et al.[30] have presented their results of the semi-automated segmentation of liver metastases treated by radiofrequency ablation. The segmentation algorithm[31] used in their study consists of six steps where a three-dimensional region growing and morphologic operations, like erosion and dilation are performed. Besides, the user needs to draw a diameter across the lesion, or, for smaller lesions, provide a single click inside the lesion as initialization. Weihusen et al.[32] have introduced a workflow oriented software support for image guided RFA of focal liver malignancies, in which they also segment coagulation necrosis in the (post-interventional) control scans after the ablation. Thereto, the user has to provide a seed point inside the ablation zone which starts a morphology-based region growing algorithm proposed by Kuhnigk et al.[33]. Afterwards, the segmentation result can be corrected towards more "irregular" or "roundish" geometry by manual interaction. Bricault et al.[34] have presented their preliminary results of a 3D shape-based analysis of CT Scans for the detection of local recurrence of liver metastases after RFA treatment. For that purpose, they applied a semi-automated 3D segmentation process that uses a "tagged" watershed algorithm. The semi-automated segmentation, on average, takes about 4 minutes and the minimum required user interaction includes two mouse clicks: one in the ablated tumor and one in the surrounding liver parenchyma. Another volumetric evaluation study of the variability of the size

 

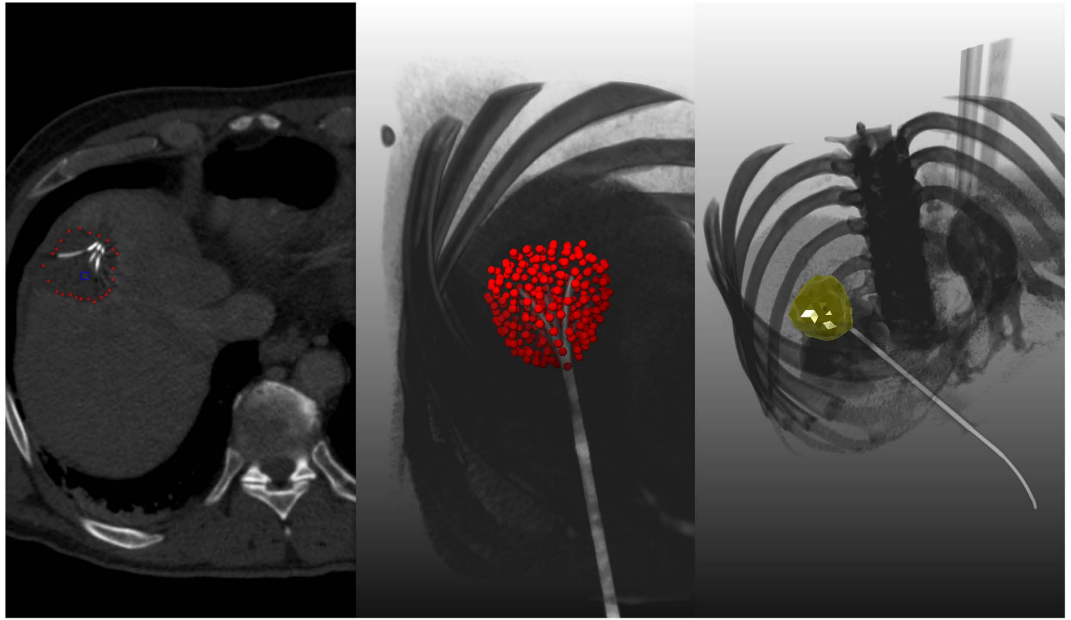

**Figure 3. This image shows overall three screenshots of an interactive segmentation result for a postinterventional CT acquisition: a screenshot of an axial plane on the left side and two 3D screenshots on the next two images to the right.** The red dots display the segmentation result in the two images on the left, besides the axial plane contains the user-defined seed point (blue) where the interactive segmentation has been stopped. Finally, the rightmost screenshot includes a closed surface (green) of the interactive segmentation result of the ablation zone, which has been generated from the red dots shown in the middle screenshot. From the closed surface on the other hand, a solid mask can be generated, which is used to determine the Dice Similarity Coefficient (DSC) if compared with a pure manual slice-by-slice expert volumetry. Note: for the native scan please see Fig. 2.

and the shape of necrosis induced by RFA in human livers has been presented by Stippel *et al.*[35]. The volumetric evaluation was performed with the software package VA 40C from Siemens on a Leonardo workstation. For the stepwise segmentation of the ablation-induced lesions, a region of interest had to be defined in each slice by manually tracking the approximate borders of the lesion. Afterwards, the precise border of the lesion was determined by a filter algorithm (provided by the software package), which is based on density differences between the ablated tissue and the liver tissue. As an enclosure to this section, we want to point the interested reader to an overview publication about computer-assisted planning, intervention and assessment of liver tumor ablation from Schumann *et al.*[36]. To the best of our knowledge, there is no work that has studied the semi-automatic 3D segmentation of post-interventional RF ablation zones, especially with clinical data that has still the ablation needles in place. Moreover, we are the first to provide a unique dataset collection from the clinical routine of post-interventional RFA cases (with and without ablation needles still in place) to the research community.

The contributions are organized as follows: At first, the Results section presents the outcomes of our experiments. Then the Discussion section concludes our contribution and outlines areas for future research. Finally, the Materials and Methods section presents the details of the proposed segmentation algorithm and online resources where anonymized (medical) data can be found.

## Results

The proposed interactive segmentation algorithm has been implemented as a C++ module (Visual Studio 2008) for the medical prototyping platform MeVisLab (www.mevislab.de, Version 2.3). The computation of the segmentation result, including the graph construction from the current user-defined seed point position and the optimal min-cut calculation could be performed within one second (on a laptop with Intel Core i5–750 CPU, $4 \times 2.66$ GHz, 8 GB RAM, Windows 7 Professional x64 Version with Service Pack 1 installed). This enables real-time feedback of our algorithm presented to the user. This immediate response and feedback of the segmentation allows user guidance of the algorithm to a satisfying outcome.

Figure 3 presents a semi-automatic segmentation result of a post-interventional ablation zone for visual inspection. As the CT data has been acquired immediately after the treatment, the needle used for the ablation is still in place and therefore visible in the scan. The left image shows the axial slice with a user-defined seed point (blue) and the red dots are the border of the segmentation in this slice. The red dots represent the last nodes of the graph that are still bound to the source *s* after the min-cut. The image in the middle presents the segmentation result in 3D. Again, the red dots show the last nodes of the graph which are still connected to the source after applying the min-cut. Finally, the rightmost image visualizes a closed surface (green) that has been generated from the graph's nodes representing the segmentation result. Afterwards, this closed surface is used to generate a solid mask for the calculation of the Dice Similarity Coefficient (DSC)[37] with the pure manual slice-by-slice segmentations.

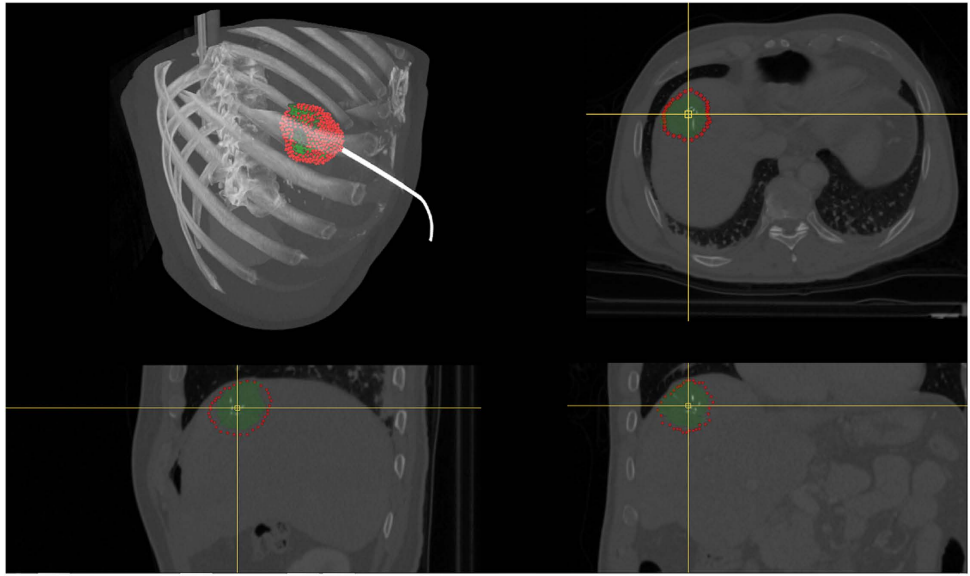

**Figure 4. This screenshots present a direct comparison between a pure manual segmentation (green) and a semi-automatic/interactive segmentation (red).** Therefore, the three-dimensional masks of both segmentations (manual/interactive) have been merged and placed within the original dataset at the location of the ablation zone (upper left window). Easily recognizable is the bright stick pointing to the masks, which is the shaft of the RFA needle. The remaining three windows show the planes where the user-defined seed point (yellow cross) has been placed for interactive segmentation result, with the axial plane in the upper right windows, the sagittal plane in the lower left window and the coronal plane in the lower right window. Note: for the native scan please see Fig. 2.

Figure 4 displays a comparison of a manual (green) and an automatic (red) segmentation for visual inspection. The upper left window shows both segmentation results as 3D masks superimposed on the original dataset. The upper right, lower left and lower right windows present direct comparisons between the manual and the automatic segmentations on an axial, sagittal and coronal slice, respectively. The yellow cross points to the location of the manually placed seed point for the graph construction. The lower left window shows that the algorithm tends to an over segmentation compared to the manual counterpart. However, changing the contrast window clearly shows the reason: the algorithm adapts to the bright border around the ablation zone (Fig. 5).

Additionally, Table 1 presents the direct comparison of manual slice-by-slice segmentations from physician 1 and results of semi-automatic segmentations for twelve ablation zones using the Dice Similarity Coefficient. Table 2 presents the summary of results from Table 2 for min, max, mean $\mu$ and standard deviation $\sigma$ for the twelve ablation zones. In the same way, Table 3 presents the direct comparison of manual slice-by-slice segmentations from physician 2 and semi-automatic segmentation results for the twelve ablation zones via the Dice Similarity Coefficient, and Table 4 presents the summary of results from Table 3 for min, max, mean $\mu$ and standard deviation $\sigma$. Furthermore, Table 5 presents the direct comparison of manual slice-by-slice segmentations from physician 1 and physician 2 for the twelve ablation zones via the Dice Similarity Coefficient, and Table 6 presents the summary of results from Table 5 for min, max, mean $\mu$ and standard deviation $\sigma$. Overall the results showed that the manual delineations (inter-observer DSC 88.8) of the lesions borders gave better results than the automatic method (DSC 77.0 – 77.1) – mean DSC values and segmented volumes of both readers were used for simplicity because individual values did not differ significantly between both readers. The differences when comparing inter-observer DSC to DSC between automatic and physician 1 ($p$-value $\ll 0.01$) and DSC automatic and physician 2 ($p$-value $\ll 0.01$) where both statistically significant. The results also showed that on average the automatically segmented lesion was smaller (33.03) when compared to physician 1 (35.85) or the physician 2 (36.18). However, this difference is not statistically significant ($p$-value 0.42 for physician 1) and ($p$-value 0.30 physician 2).

Moreover, Table 7 and Table 8 present the results from Table 1, but differentiate between the cases where the RF electrodes were still visible (Table 7) and the cases where the RF electrodes have already been removed (Table 8). As Table 7 and Table 8 show, there is no significant differences between the cases with and without RF electrodes visible in the images. In more detail, the DSC values between readers were significantly higher ($p < 0.01$) than those between automatic and manual processing: 88.8% vs. 77.0%, also independent of whether the needle was still included (86.8% vs. 75.9%, $p < 0.05$) in the dataset or not (90.9% vs. 78.1%, $p < 0.05$), respectively. Segmented volumes appeared to be smaller with automatic processing than for readers, but mean differences were not significant: 33.03 ml vs. 36.02 ml (p = 0.308). This also held for cases with the needle included (25.76 ml vs. 25.93 ml, p = 0.917) and without (40.30 ml vs. 46.10 ml, p = 0.249). The mean DSC value of both readers appeared to be smaller when the needle was present (75.9% vs. 78.1% without, p = 0.423) but the difference was not significant. In contrast, the inter-observer DSC was significantly higher when the needle was not present (90.9% vs. 86.8%, p = 0.025). Statistical differences in DSC values and segmentation volumes between methods were analyzed with Wilcoxon signed-rank tests, differences between both subgroups (with and without needle present) with a Mann-Whitney test. All analyses were done with SPSS (Version 20, IBM) using a level of significance of 0.05.

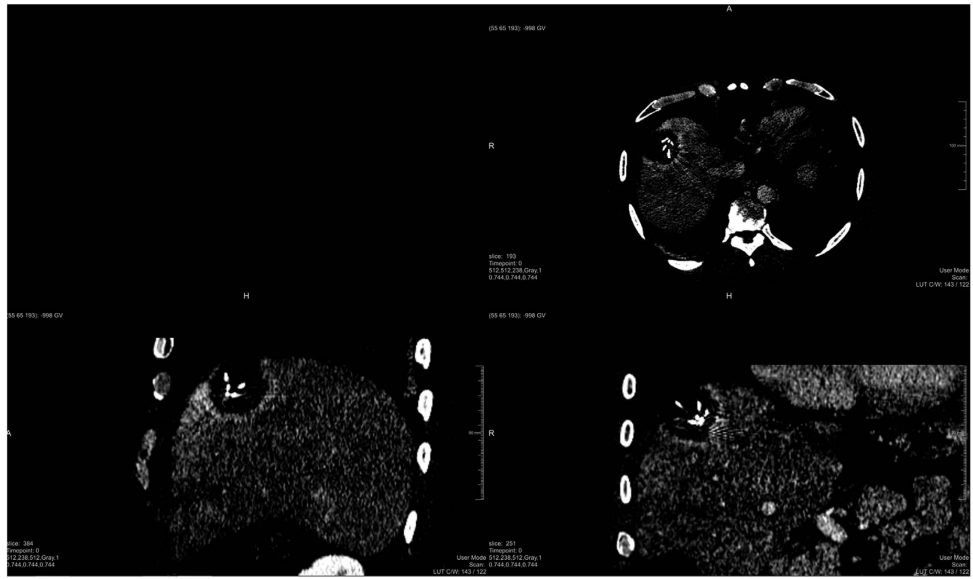

**Figure 5. Extreme windowing setting for the acquisition and planes from Fig. 4, making a bright border around the ablation zone recognizable.** Note: for the native scan with an appropriate windowing please see Fig. 2.

| No. | volume of AZ (mm³) | | number of voxels | | DSC (%) |
| --- | --- | --- | --- | --- | --- |
| | manual | automatic | manual | automatic | |
| 1 | 30592.2 | 38900.5 | 55246 | 70208 | 81.98 |
| 2 | 18059.4 | 27123.2 | 32785 | 49175 | 71.78 |
| 3 | 13785.6 | 16672.5 | 8255 | 9988 | 72.75 |
| 4 | 32216.4 | 24520.8 | 18870 | 14348 | 76.36 |
| 5 | 33560.7 | 22190.1 | 58742 | 38852 | 72.49 |
| 6 | 40618.7 | 36919.6 | 28974 | 26319 | 82.23 |
| 7 | 122617 | 104022 | 70806 | 60113 | 83.43 |
| 8 | 12438.4 | 18105.4 | 21328 | 31069 | 73.05 |
| 9 | 9950.5 | 6253.58 | 5866 | 3689 | 71.94 |
| 10 | 49762.4 | 50814.2 | 34118 | 34835 | 83.53 |
| 11 | 40484 | 25717.6 | 26122 | 16593 | 75.85 |
| 12 | 26151.3 | 25146.7 | 14426 | 13887 | 78.28 |

**Table 1. Direct comparison of manual slice-by-slice segmentations from physician 1 and semi-automatic segmentation results for twelve ablation zones (AZ) via the Dice Similarity Coefficient (DSC).**

| | volume of AZ (cm³) | | number of voxels | | DSC (%) |
| --- | --- | --- | --- | --- | --- |
| | manual | automatic | manual | automatic | |
| min | 9.95 | 6.25 | 5866 | 3689 | 71.78 |
| max | 122.62 | 104.02 | 70806 | 70208 | 83.53 |
| $\mu \pm \sigma$ | 35.85 ± 30.04 | 33.03 ± 25.14 | 31294.83 | 30756.33 | 76.97 ± 4.73 |

**Table 2. Summary of results from Table 1: min, max, mean $\mu$ and standard deviation $\sigma$ for twelve ablation zones.**

| No. | volume of AZ (mm³) | | number of voxels | | DSC (%) |
|---|---|---|---|---|---|
| | manual | automatic | manual | automatic | |
| 1 | 31114.4 | 38900.5 | 56189 | 70208 | 80.69 |
| 2 | 16629.4 | 27123.2 | 30189 | 49175 | 68.9 |
| 3 | 14159.7 | 16672.5 | 8479 | 9988 | 75.25 |
| 4 | 32919.8 | 24520.8 | 19282 | 14348 | 76.15 |
| 5 | 31437.6 | 22190.1 | 55026 | 38852 | 74.23 |
| 6 | 42929 | 36919.6 | 30622 | 26319 | 81.19 |
| 7 | 117694 | 104022 | 67963 | 60113 | 85.19 |
| 8 | 12651.2 | 18105.4 | 21693 | 31069 | 73.71 |
| 9 | 11098.9 | 6253.58 | 6543 | 3689 | 68.14 |
| 10 | 50345.8 | 50814.2 | 34518 | 34835 | 85.29 |
| 11 | 42635.1 | 25717.6 | 27510 | 16593 | 73.73 |
| 12 | 30576.3 | 25146.7 | 16867 | 13887 | 82.45 |

**Table 3. Direct comparison of manual slice-by-slice segmentations from physician 2 and semi-automatic segmentation results for twelve ablation zones (AZ) via the Dice Similarity Coefficient (DSC).**

| | volume of AZ (cm³) | | number of voxels | | DSC (%) |
|---|---|---|---|---|---|
| | manual | automatic | manual | automatic | |
| min | 11.1 | 6.25 | 6543 | 3689 | 68.14 |
| max | 117.69 | 104.02 | 67963 | 70208 | 85.29 |
| $\mu \pm \sigma$ | $36.18 \pm 28.72$ | $33.03 \pm 25.14$ | 31240.08 | 30756.33 | $77.08 \pm 5.83$ |

**Table 4. Summary of results from Table 3: min, max, mean $\mu$ and standard deviation $\sigma$ for twelve ablation zones.**

| No. | volume of AZ (mm³) | | number of voxels | | DSC (%) |
|---|---|---|---|---|---|
| | manual 1 | manual 2 | manual 1 | manual 2 | |
| 1 | 30592.2 | 31114.4 | 55246 | 56189 | 90.74 |
| 2 | 18059.4 | 16629.4 | 32785 | 30189 | 82.35 |
| 3 | 13785.6 | 14159.7 | 8255 | 8479 | 86.4 |
| 4 | 32216.4 | 32919.8 | 18870 | 19282 | 89.29 |
| 5 | 33560.7 | 31437.6 | 58742 | 55026 | 86.7 |
| 6 | 40618.7 | 42929 | 28974 | 30622 | 92.61 |
| 7 | 122617 | 117694 | 70806 | 67963 | 92.61 |
| 8 | 12438.4 | 12651.2 | 21328 | 21693 | 87.54 |
| 9 | 9950.5 | 11098.9 | 5866 | 6543 | 88.81 |
| 10 | 49762.4 | 50345.8 | 34118 | 34518 | 91.22 |
| 11 | 40484 | 42635.1 | 26122 | 27510 | 92.48 |
| 12 | 26151.3 | 30576.3 | 14426 | 16867 | 85.02 |

**Table 5. Direct comparison of manual slice-by-slice segmentations from physician 1 and physician 2 for twelve ablation zones (AZ) via the Dice Similarity Coefficient (DSC).**

| | volume of AZ (cm³) | | number of voxels | | |
|---|---|---|---|---|---|
| | manual 1 | manual 2 | manual 1 | manual 2 | DSC (%) |
| min | 9.95 | 11.1 | 5866 | 6543 | 82.35 |
| max | 122.62 | 117.69 | 70806 | 67963 | 92.61 |
| $\mu \pm \sigma$ | 35.85 ± 30.04 | 36.18 ± 28.72 | 31294.83 | 31240.08 | 88.81 ± 3.3 |

**Table 6. Summary of results from Table 5: min, max, mean $\mu$ and standard deviation $\sigma$ for twelve ablation zones.**

| | volume of AZ (cm³) | | number of voxels | | |
|---|---|---|---|---|---|
| | manual | automatic | manual | automatic | DSC (%) |
| min | 13.79 | 16.67 | 8255 | 9988 | 71.78 |
| max | 33.56 | 138.9 | 58742 | 70208 | 81.98 |
| $\mu \pm \sigma$ | 25.73 ± 8.11 | 25.76 ± 7.37 | 31387.33 | 32743 | 75.61 ± 4.02 |

**Table 7. Summary of results from Table 1 presenting only the cases where the RFA needle has been still in place and thus visible in the images (cases 1–5 and case 12): min, max, mean $\mu$ and standard deviation $\sigma$ for six ablation zones.**

| | volume of AZ (cm³) | | number of voxels | | |
|---|---|---|---|---|---|
| | manual | automatic | manual | automatic | DSC (%) |
| min | 9.95 | 6.25 | 5866 | 3689 | 71.94 |
| max | 122.26 | 104.02 | 70806 | 60113 | 83.53 |
| $\mu \pm \sigma$ | 45.98 ± 40.91 | 40.31 ± 34.78 | 31202.33 | 28769,67 | 78.34 ± 5.35 |

**Table 8. Summary of results from Table 1 presenting only the cases where the RFA needles has already been removed (cases 6–11): min, max, mean $\mu$ and standard deviation $\sigma$ for six ablation zones.**

The main difference between the automatic and manual segmentation is the computation time. Using the proposed automatic tool the segmentation was done in few seconds whereas the manual segmentation took between 48 s and 8 min 16 s (mean 3 min 13 s). Note: some initial results have been presented and discussed at the 100[th] annual meeting of the Radiological Society of North America (RSNA) in December 2014[38] and a German computer science workshop (BVM) in March 2015[39]. However, at the RSNA meeting we only showed an initial statistic summary of the segmentation outcome and at the BVM workshop we presented a summarized description of the interactive algorithm in German. Moreover, all statistical results and a precise description of the methods are only presented in full details (and in English) within this manuscript.

## Discussion

RFA of liver tumors induces areas of tissue necrosis, which can be visualized reliably in contrast enhanced CT. In this study, an interactive segmentation algorithm was applied to datasets of routine control CT scans after RFA of liver cancer for semi-automatic determination of thermally induced lesions. The segmentation accuracy was found to be sufficient for lesion segmentation in most of the cases although the manual segmentation still provided the best segmentation accuracy. The main advantage of the proposed tool over manual segmentation is its speed, which makes it an appealing alternative application for physicians.

As discussed, RFA can stand as a minimally invasive alternative to open surgery and might also be suitable for patients who are inoperable or refuse surgery. In RFA, post-interventional imaging is regularly performed to document the success of the treatment. When the interventional radiologist presumes that continuing of the therapy might be necessary, the RFA needle may still reside in the target organ during image acquisition. Due to hardening artefacts, image quality can be compromised significantly. However, these datasets have also been segmented and included into the study. For the evaluation of the contouring algorithm, manual slice-by-slice segmentations have been performed by two radiological experts, which enabled the DSC calculation between the manual and the semi-automatic segmentation outcome. Furthermore, we will provide unique datasets – including the manual segmentations – to the community for their own research purposes. In summary, the achieved research highlights of the presented work are:

- Applying an interactive segmentation algorithm to RFA datasets from the clinical routine;
- Incorporating intraprocedural patient acquisitions with and without RFA needles into the evaluation set;
- Performing pure manual slice-by-slice segmentations by radiological experts for quantitative and qualitative evaluation;
- Calculation of the Dice Similarity Coefficient (DSC) for statistical validation of the presented segmentation algorithm;
- Providing the anonymized RFA data collection with the corresponding manual expert segmentation to the research community.

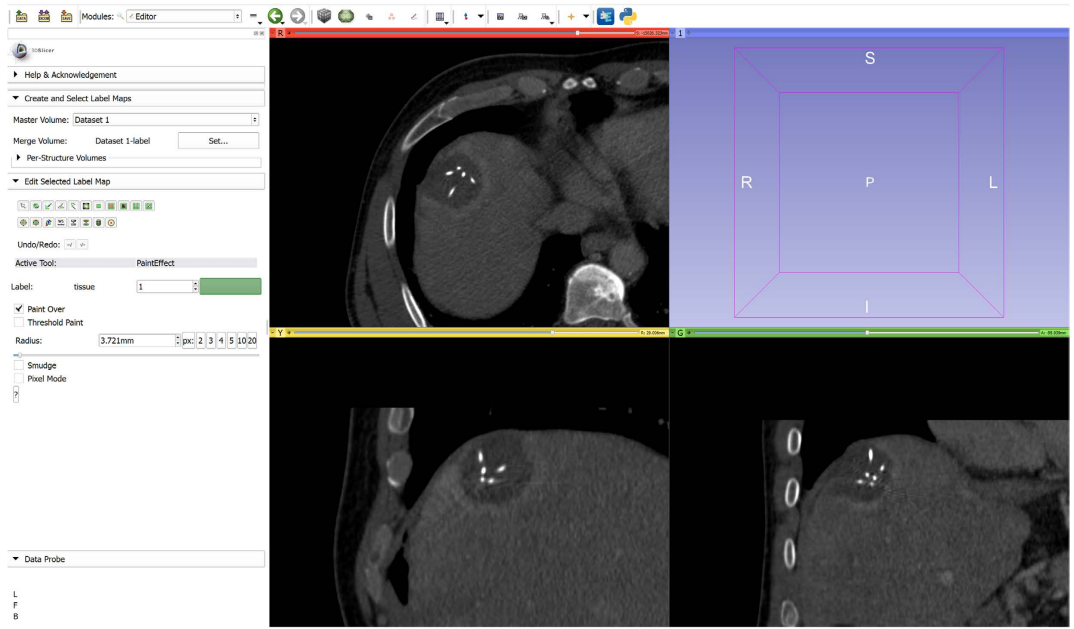

**Figure 6. Postinterventional CT scan of an RF ablation with the ablation needle still in place loaded into 3D Slicer.** On the left side is the Editor module which also contains the GrowCut algorithm.

For comparison of our employed method with a state-of-the-art segmentation approach we used GrowCut[40], which is freely available under the medical platform 3D Slicer (Fig. 6). The implementation is very user friendly because it does not require any precise parameter setting, rather the user initializes the method by marking areas in the image with simple strokes (Fig. 7). Furthermore, we had good experiences with certain types of brain tumors (Glioblastoma multiforme (GBM)[41] and pituitary adenomas[42]). However, we tested the GrowCut algorithm with our RFA datasets and especially the cases where the needles are still in place – and thus visible within the images – caused massive problems (Fig. 8). What happens is, that GrowCut leaks along the needles, because it cannot handle the large gray value differences between the ablation zone (dark) and the RFA needle (bright). Overall we tested four cases from our datasets with GrowCut: two cases with the needles still in place and two cases without needles. For the two cases with needle in place, we could only archive a DSC of 50.64% (for the case presented in Fig. 8) and a DSC of 50.28% for the second case. As mentioned before, for both cases GrowCut leaks along the RFA needles, which explains the low Dice Scores. However, for the cases without needle in place we could archive Dice Scores of 80.3% and 76.29%. Here, in contrast to the cases with needle, the leaking did not occur, which resulted in acceptable segmentation results. But still, our approach could achieve higher Dice Scores of 82.23% and 83.43%, respectively. Moreover, the initialization of GrowCut (marking parts of the lesion and the background) takes between 30 and 60 seconds for a trained user, in contrast to our method that needs only a single seed point. In addition, the user has to wait several seconds for the GrowCut segmentation result (on the same PC we measured about 10 seconds for the interactive method), whereas our method provides the segmentation result immediately in real-time. This makes a refinement much more convenient. Note: the sharp edges of the GrowCut result (green) in the rightmost image of Fig. 8 occur, because the Slicer implementation restricts the segmentation area with a bounding box. The size of the bounding box depends on the initialization of the user and avoid GrowCut to use the whole image or volume for the automatic segmentation process. In addition to GrowCut we also tested and evaluated existing (implemented) segmentation approaches from other medical plat-forms, like The Medical Imaging Interaction Toolkit (MITK, www.mitk.org) and MeVisLab (see Results section) with our data. MITK, developed by the German Cancer Research Center (DKFZ) in Heidelberg, Germany, combines also the Insight Toolkit (ITK, www.itk.org) and the Visualization Toolkit (VTK, www.vtk.org) with an application framework. Amongst others, we applied the Fast Marching 3D imple-mentation from MITK (version MITK-2015.05) to our data, but it could also not handle the extreme gray value differences between the needle (bright) and the ablation zone (dark). However, in contrast to GrowCut the Fast Marching algorithm did not leak along the needle, rather it excluded the needle. Also placing additional seed points direct on the needle and thus providing the algorithm the information about the bright parts did not lead to a satisficing segmentation result. The DSCs for the two cases with the needles still in place were 51.94% and 30.01%. But for the none needle cases we could achieve better results, which resulted in Dice Scores of 73.56% and 63.93% compared with the manual segmentations. However, beside the manual seed points that had to be placed for the approach to run, there are several segmentation parameters (Sigma, Alpha, Beta, Stopping value and Threshold), which make it on one hand quite difficult to find a good parameter setting. On the other hand the segmentation results could probably be better with a more precise parameter tuning. We also tried the Region Growing 3D from MITK but for all cases (needle and none needle) the approach leaked into the surrounding structure. However, if the seed point was placed directly on the needle within the image, the whole needle could be segmented quite well (due to the bright values of the needle). As Level-Set segmentation method we used the itkGeodesicActiveContourLevelSetImageFilter-module from the current MeVisLab version (MeVisLab 2.7, 18–06–2015), which wraps the ITK class GeodesicActiveContourLevelSetImageFilter[43]. Beside an arbitrary amount of user-defined seed points several parameters can be tuned for the segmentation, but a parameter change and re-segmentation took over two minutes, which made it quite time-consuming to find a good setting (note: the same laptop – described in the result section and used for the presented interactive segmentation – was also used for the MITK and MeVisLab segmentation approaches). However, the first case (with needle) showed a DSC of 60.98% and leaked along the needle. The second case with needle had only a DSC of 22.56% and leaked along the needle and into the surrounding structures. This happened because the gray values of the ablation zone were more similar to the surrounding tissue for this case. For the none needle cases we achieved Dice Scores of 45.63% and 70.18% and the lower DSC resulted, because we could only archive an under or an over segmentation of the ablation zone (the DSC presented here is for the under segmentation).

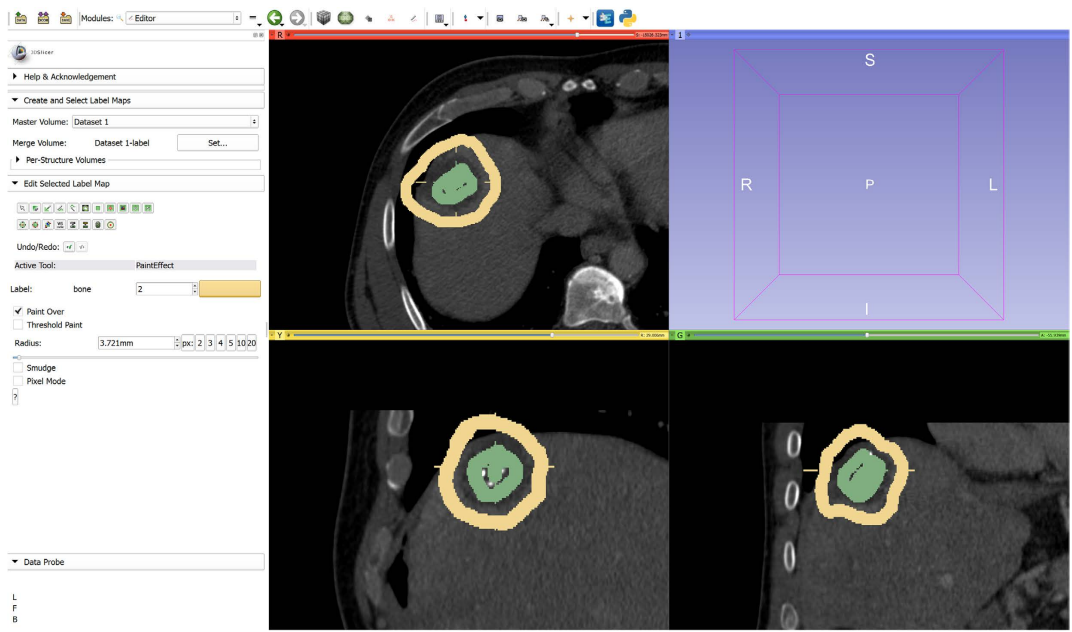

**Figure 7. GrowCut initialization for the segmentation of the RF ablation zone: the ablated zone is marked in green and the background is marked in yellow on three 2D slices, respectively.**

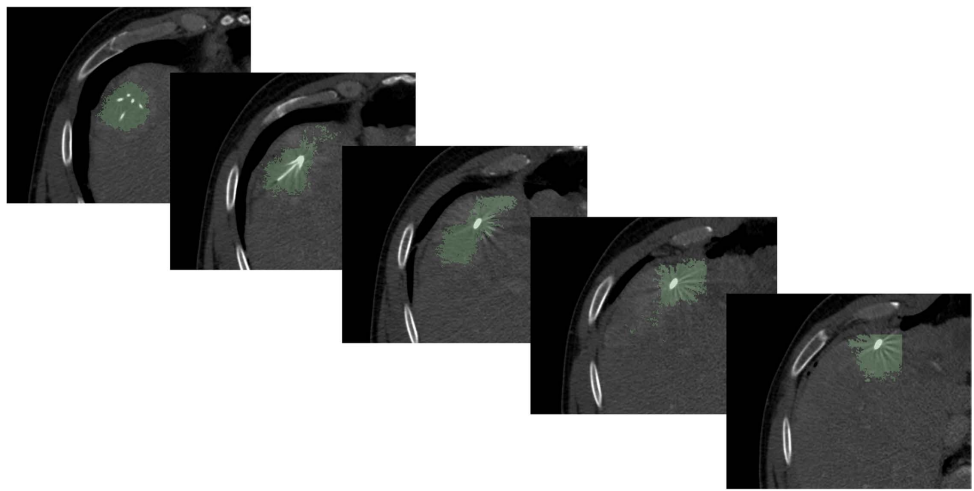

**Figure 8. GrowCut segmentation result (green) for the initialization from Fig. 7.** The GrowCut segmentation leaks along the RFA needle, because it cannot handle the large gray value differences between the ablation zone (dark) and the needle (bright). Note: the sharp edges of the segmentation result in the rightmost image occur because the GrowCut implementation in 3D Slicer automatically restricts the segmentation area with a bounding box that depends on the user initialization.

There are several areas for future work: we plan to integrate the interactive segmentation algorithm into a medical application framework for supporting ablation therapy. This is currently being developed within a project funded by the European FP7 program. In particular, our semi-automatic algorithm is targeted for the segmentation of difficult cases where an automatic segmentation fails. Furthermore, we also plan to provide the datasets acquired during the duration of the project over the next years and want to investigate the use for lesion tracking after one or several RFA interventions. Moreover, we hope to improve the segmentation outcomes with hardware accelerated medical image processing. One parameter of our approach, is the amount of nodes that are used for the graph construction. These result from the number of rays (sent out from the user-defined seed point) and the sampled nodes along every ray. As described in the Material and Methods section we use the surface points from a polyhedron to enable a faster calculation of the rays, which means we rely on recursively refined polyhedra, like 12, 32, 92, 272, 812, 2432 vertices. At present, we utilize maximal 812 surface points and maximal 40 points per ray to still allow an interactive real-time feedback segmentation for the user. A greater number of nodes would cause too much time delay between the single segmentation calculations and break the real-time character of the presented approach. However, we would like to apply a greater number of rays – e.g. 2432 – in the future and expect that this leads to even more precise segmentation results. However, the mincut calculation is currently the bottle neck which does not permit a greater number of nodes current standard hardware configuration, but a solution may be the execution on a GPU.

## Material and Methods

**Data Acquisition.**   For this retrospective study we used twelve intraprocedural datasets from ten patients, who underwent a radiofrequency ablation of a liver tumor. All datasets had a matrix size of $512 \times 512$ in x- and y-direction, and the number of slices in z direction ranged from 52 to 232. The slice thickness was either 1 or 2 mm, and the pixel spacing ranged from 0.679 to 0.777 with spacing between the slices of 1 to 3 mm. In six datasets (cases 1–5 and cases 12), the ablation needle was still remaining in the liver. All datasets have been acquired on multislice CT scanner (Philips Brilliance or Mx8000, Philips Healthcare, Netherlands). The data collection, analysis and a future publication, was approved by the Institutional Review Board (IRB) of the Medical University of Leipzig, Germany (reference number: 381–14–15122014). The methods were carried out in accordance with the approved guidelines. Hence, the RFA data will soon be freely available for own research purposes from the official webpage of the European Project ClinicImppact (please cite this publication if you use any of these in your own work): www.clinicimpact.eu.

Note: since this is a European Project scheduled at least for three years until 2017, we will add data from several clinical partners around Europe. Furthermore, even more ablation data from a comprehensive pig study can be found on the webpage of the European Project GoSmart[44]: www.gosmart-project.eu.

**Manual Segmentation.**   To generate the Ground Truth of the ablation zones, we set up a segmentation framework under MeVisLab which provided classic contouring capabilities. This allowed the physicians to manually outline RFA lesions in the datasets slice-by-slice without any algorithmic support to avoid distortions. Afterwards, the single contours were voxelized (Fig. 9) and then merged to a 3D mask representing the ablation zone. These 3D masks have been used for comparison and quantitative evaluation with the semi-automatic segmentation results.

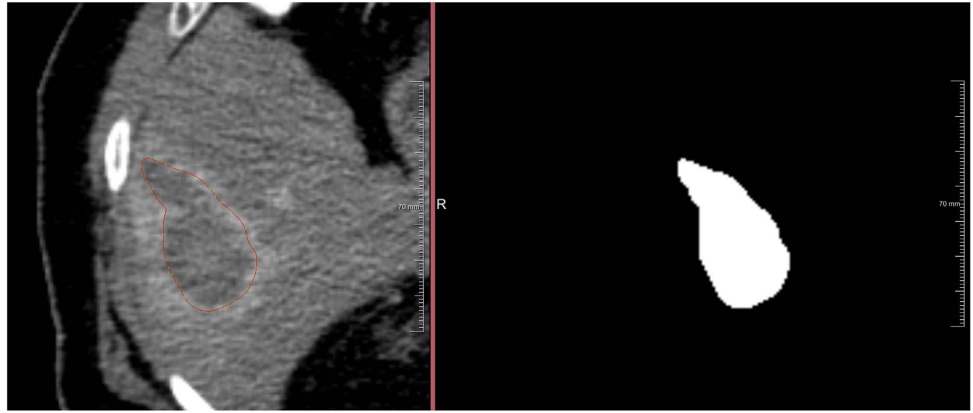

**Figure 9. The left side shows a manually outlined ablation zone (red) on a single 2D slice and the right side presents the corresponding voxelized mask (white).** All voxelized 2D slices are merged to one 3D mask representing the whole ablation zone. This manual segmentation is used to calculate the Dice Similarity Coefficient (DSC) with the segmentation result from the algorithm.

**Evaluation Metric.** As an evaluation metric we used the Dice Similarity Coefficient (DSC)[37]. The DSC is a common metric widely used in medical image analysis where the agreement between two binary volumes $M$ and $S$ is calculated:

$$DSC = \frac{2 \cdot V\,(M \cap S)}{V\,(M)\,+\,V\,(S)} \tag{1}$$

Here, $M$ and $S$ are the binary masks from the manual ($M$) and the semi-automatic ($S$) segmentations, $V(\cdot)$ denotes the volume (in mm³) and $\cap$ denotes the intersection. We computed the volume by counting the number of voxels and multiplying them with the physical size of voxels. In addition to the DSC, we measured the time it took an experienced radiologist in order to manually outline the ablation zones and compared it with the computation time of our semi-automatic segmentation procedure.

**Semi-automatic Segmentations.** The semi-automatic segmentation algorithm uses a spherical template to set up a three-dimensional graph $G(V,E)$ around the ablation zone[45]. Overall, the graph consists of nodes $n \in V$ and edges $e \in E$ connecting these nodes. Thereby, the nodes are sampled along rays whose origin resides at a user-defined seed point while their direction points towards the surface of a polyhedron enclosing the seed point[46]. In addition we use two virtual nodes: a source $s$ and a sink $t$ to construct the graph. After graph construction, the segmentation result is calculated by dividing the graph into two sets of nodes via a Min-Cut/Max-Flow algorithm[47]. As a result, one set of nodes represents the ablation zone (foreground) and is bound to one of the virtual nodes, e.g. the sink $t$. The other set represents the surrounding structures (background) and is bound to the other virtual node (in this case the source $s$). The energy function of the graph cuts follows the Gibbs model[48] and the cost function relates to the publication of Li *et al.*[49] where you need a fixed average gray value of the region to calculate the single weights. Moreover, the approach was designed to segment longish structures like the aorta[50,51] and needed a centerline as input, which makes it not applicable for an interactive real-time approach. In contrast, our approach needs only one seed point that can also be used to derive the average gray value on the fly during the segmentation and thus makes it gray value independent. Furthermore, this means that it can also handle cases with different average ablation values. Amongst others, the basic segmentation scheme has already been successfully applied to pituitary adenomas[52] and prostate central glands (PCG)[53]. The underlying workflow is shown in Fig. 10 for a post-interventional CT scan of a patient whose liver tumor has been treated with an RFA: a spherical template (blue, leftmost image) has been applied to set up the three-dimensional graph $G(V,E)$ – consisting of nodes and edges – in the second image from the left. This graph $G$ is automatically constructed at a seed point position within the image (here indicated by several 2D slices) – note: in general the graph is not visible to the user during the segmentation process, rather it is automatically constructed in the background. In this example the RFA needle is still visible within the image (bright parts inside the liver) and especially in the second and third image the characteristic umbrella shape of the fully expanded RFA needle is noticeable. Afterwards, as above already mentioned, the graph is automatically divided by the Min-Cut/Max-Flow algorithm into two disjoint sets of nodes: one set representing the ablation zone and the other one the surrounding background. However, the transition between these two sets – or in other words the last nodes of every ray that are still bound to the user-defined seed point – is the actual segmentation result that is displayed to the user in the rightmost image of Fig. 10 (red). Additionally, note that these red dots in the rightmost

 

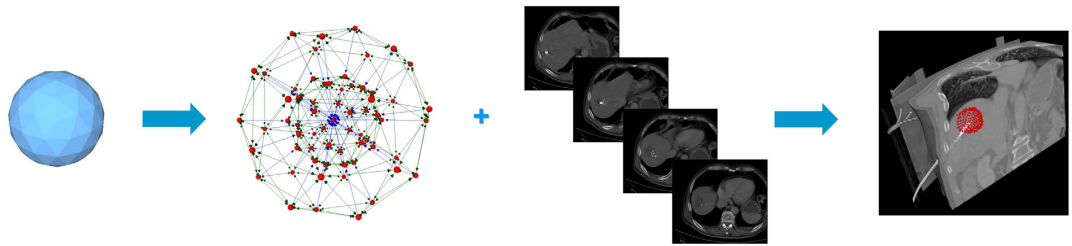

**Figure 10. Overall workflow of the RF ablation zone segmentation: a sphere (left) is used to construct a graph (second image from the left).** The graph is constructed (not visible to the user) at the user-defined seed point position within the image (third image from the left). Finally, the segmentation result (red) corresponding to the seed point is shown to the user (rightmost image).

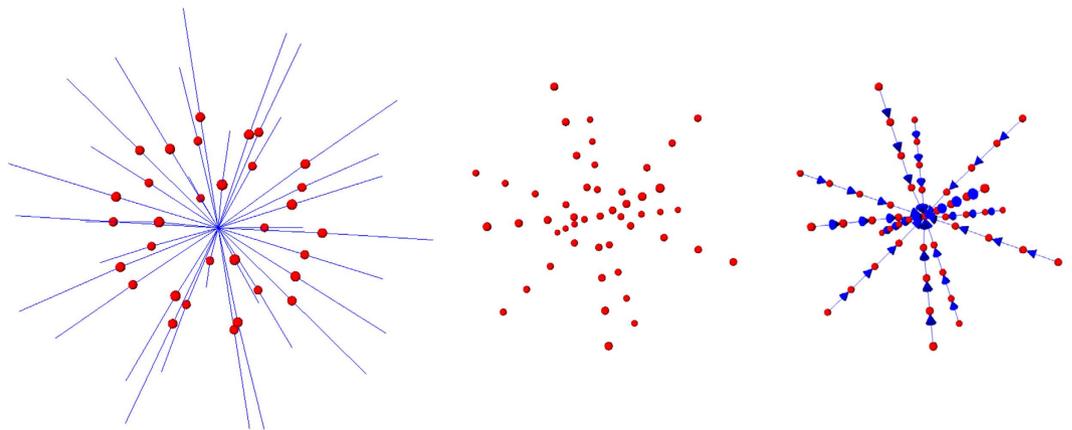

**Figure 11. More detailed graph construction for intra-edges of the segmentation approach.** Left image: in a first step rays (blue) a send through the surface points (red) of a polyhedron. Middle image: the graph's nodes are sampled along the rays from the previous image. Right image: the intra-edges (blue arrows) are constructed between the sampled nodes (red).

image (representing the segmentation result) were actually sampled nodes during the graph construction in the beginning.

In more detail, the whole set of edges $E$ (of a constructed graph $G$) consists of edges that connect the sampled nodes to the source and the sink, and edges that establish connections between two sampled nodes. However, in a first step rays (leftmost image of Fig. 11, blue lines) are send out from the user-defined seed point through the surface points of an polyhedron (leftmost image of Fig. 11, red dots). Afterwards, the graphs nodes are sampled along these rays between the user-defined seed point and the surface points of the polyhedron (image in the middle of Fig. 11, red dots). Then, the so called intra-edges between the nodes (red) along one ray are constructed (rightmost image of Fig. 11, blue arrows). These edges ensure that all nodes below a segmented surface in the graph are included to form a closed set, or in other words, the interior of the object (the ablation zone) is separated from the exterior (the surrounding background) in the data. Next, the inter-edges of the graph are constructed. These edges establish (a) connections between nodes that have been sampled along different rays and (b) connections from the sampled nodes to the source and the sink. Thereby, the intra-edges (a) constrain the set of possible segmentations and enforce the smoothness of the segmentation result over an integer parameter $\Delta r$. The larger this parameter is, the larger is also the number of possible segmentations and a value of zero enforces the segmentation result to be a sphere. Figure 12 exemplarily shows the inter-edges (between the nodes sampled along three rays) for three different $\Delta r$ values: zero (leftmost image), one (image in the middle) and two (rightmost image). Supplementary, Fig. 13 demonstrates how the different $\Delta r$ values influence the segmentation outcome for two adjacent rays and their sampled nodes. For the leftmost image of Fig. 13, a $\Delta r$ value of zero was chosen. Thus, only inter-edges between nodes on the same "node level" are established, which will also lead the mincut to be on the same "node level" of different rays (green); elsewise "costs" would arise by cutting the inter-edges (which will the mincut automatically avoid). Furthermore, the position (or the "node level") of the cut (in this example between the second and third nodes when counting from the bottom), depends on other factors like the gray values in the image. However, in any case the segmentation outcome will be a sphere and the "node level" of the cut determines the size of this sphere. The next three images of Fig. 13 will illustrate what happens for a $\Delta r$

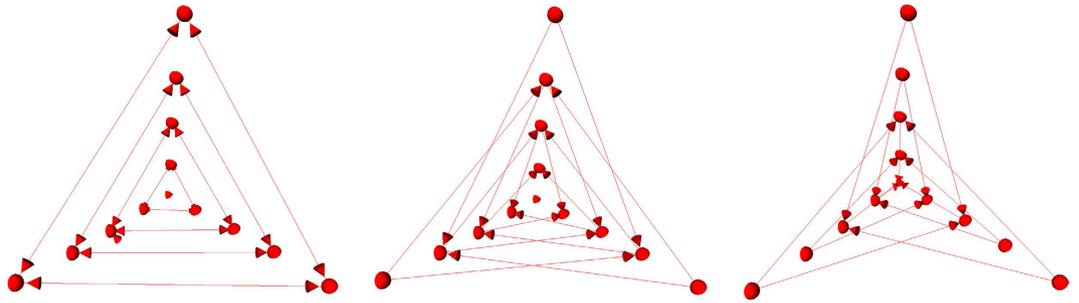

**Figure 12. Principle of the inter-edge constructions (red arrows) between nodes (red dots) that have been sampled along three different rays.** The leftmost image shows the inter-edges for a $\Delta r$ value of zero. Thus the inter-edges are all on the same "node level". The image in the middle shows the inter-edges for a $\Delta r$ value of one, resulting in edges that connect nodes from different "node levels", however, with a maximum level difference of one. Finally, the rightmost image shows the inter-edges for a $\Delta r$ value of two, connecting nodes with a "node level" distance of two. Similarly, this practice also applies for larger values of $\Delta r$, e.g. three or four and so on.

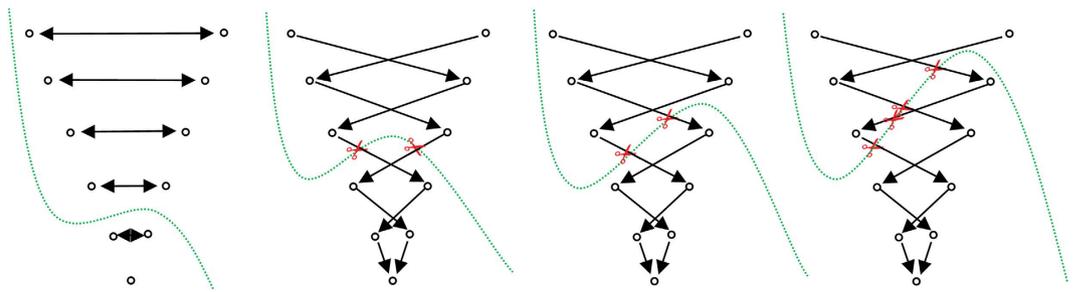

**Figure 13. This figure shall illustrate the course of action of the mincut for different $\Delta r$ values.** On the left side, the inter-edges for a $\Delta r$ value of zero have been constructed. Thus the mincut will separate all nodes on the same "node level" to avoid costs for cutting inter-edges. Note that the location of the cut (green) depends here on other factors, like the underlying gray values. Similar to the second image of Fig. 12, the inter-edges for a $\Delta r$ value have been constructed for the following three rightmost images. As you can see, the mincut has two options for cutting inter-edges (red scissors) producing the same costs: (1.) on the same "node level" (second image from the right) and (2.) cutting on different "node levels" with a node distance on one (third image from the right). However, cutting on different "node levels" with a node distance on two (or greater) will produce higher cost and therefore will automatically be avoided by the mincut algorithm as seen in the rightmost figure. The same principle applies also for larger values of $\Delta r$.

value of one. As you can see, the inter-edges have been constructed between different "node levels" equivalent to the image in the middle of Fig. 12. For a $\Delta r$ value of one, the mincut has definitely to cut two inter-edges (red scissors), regardless if the cut is on the same "node level" (second image from the right of Fig. 13) or if the cut is between different "node levels" (second image from the right of Fig. 13). However, for a $\Delta r$ value of one, this only applies if the cut also appears within a maximum "node level" distance of one. For a cut with a "node level" distance of two (or larger) and a $\Delta r$ value of one, "costs" for cutting four (or more) inter-edges would arise as shown in the rightmost image of Fig. 13 (note: the mincut algorithm will automatically avoid this cut). Accordingly, this principle applies to larger $\Delta r$ values, like 2, 3, etc. Once the connections between the nodes (sampled nodes and virtual nodes) have been established, a weight is assigned to every edge. These weights are the costs that arise when the mincut algorithm cuts through the edges. The intra-edges along one ray and the inter-edges resulting from the $\Delta r$ value are assigned a maximum value $\infty$, e.g. the maximum float value when implemented. The weights of the edges connecting the sampled nodes $v_{(x,y,z)} \in V$ with the two virtual nodes source and sink depend on the gray values of the image, namely the positions of the sampled nodes. Technically speaking, the weights depend on cost-values $c_{(x,y,z)}$ describing the absolute value of the difference of an average ablation zone value and the gray value at $v_{(x,y,z)}$. Figure 14 provides an example for nodes that have been sampled along one ray and then bound via the absolute value of the difference of an average (ablation zone) value to the source (red) and the sink (blue). Additionally, the intra-edges (1.-8.) are drafted in the Figure. As you can see the lowest cost of ninety arise when the mincut algorithm cuts through the third intra-edge (green). The following Fig. 15 emphasizes the value of the intra-edges: if

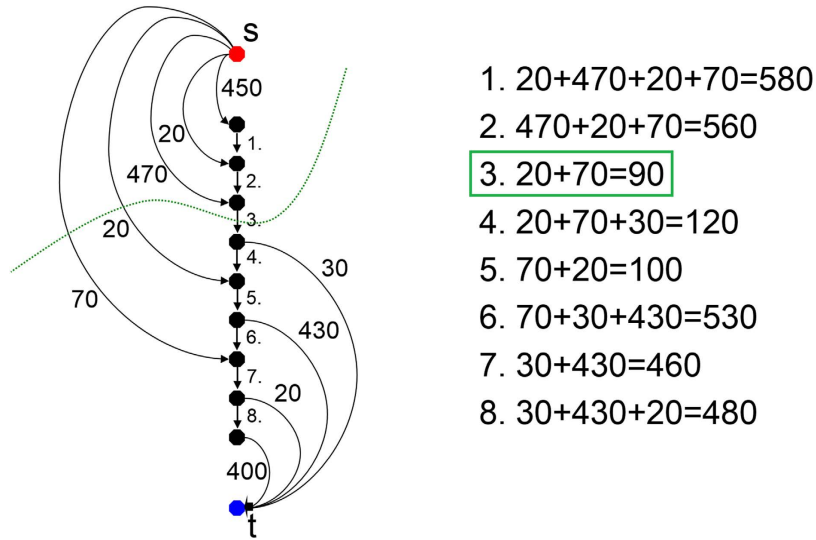

1. 20+470+20+70=580
2. 470+20+70=560
3. 20+70=90
4. 20+70+30=120
5. 70+20=100
6. 70+30+430=530
7. 30+430=460
8. 30+430+20=480

**Figure 14. This image shows several sampled nodes for a graph that has been bound by edges and weights to the source (red) and the sink (blue).** Besides, the eight intra-edges between the sampled nodes have been generated. As seen, the minimal s-t-cut cuts the third intra-edge and produces a total cost of ninety.

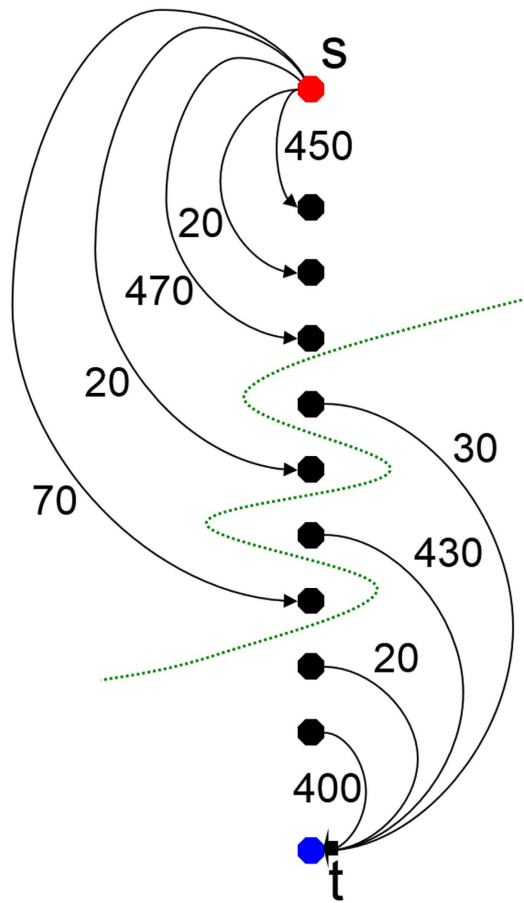

**Figure 15. This mincut example adapted from Fig. 14 discloses the benefit of the intra-edges (which have not been generated here): the s-t-cut (green) will avoid cutting the inter-edges to produce an overall "cutting" cost of zero.** Hence, this would no longer ensure that all nodes below a segmented surface in the graph are included to form a closed set.

removed from Fig. 14, the mincut could avoid cutting any edges (green) resulting a total cost of zero. However, the average ablation zone (gray) value has a huge influence on the segmentation result, but we can assume that the user places the seed point inside the ablation zone. Thus we can use this information to determine the average ablation zone value during the interactive segmentation. In addition, to avoid outliers, e.g. produced by the RFA needle (if the needle is still in place and therefore visible within the image), we integrate over a small area of about one $cm^3$ around the current position of the user-defined seed point. In doing so, we handle situations, where the user places the seed point at a position of the RFA needle within the image and the corresponding gray value is much too bright/large to be used as an average ablation zone value. This strategy also enables to automatically handle different image acquisitions and makes the algorithm kind of disease invariant and scan independent (for example against possible inhomogeneities within ablation zones). But the greatest advantage is still the automatic handling of scans with and without RFA needles visible within the medical image.

Following, this basic segmentation scheme was later turned into an interactive real-time approach[54] and for this study it was enhanced by an additional refinement option[55,56,57]. For starting the interactive segmentation process, the user places a seed point roughly in the middle of the ablation zone on a 2D slice. From this seed point we construct the graph and automatically calculate and display the segmentation result for the user. The user now has the option to drag the seed point around to interactively generate new segmentation results depending on the current seed point positions. Additionally, the user can interrupt the dragging of the seed point by releasing the mouse button and add an arbitrary amount of seed points on the border of the ablation zone. Thus, the algorithm gets supporting input about the location of the border and steers the behavior of the min-cut. However, the user can always come back to the initial seed point and start dragging it around in the image again (note: additionally placed seed points will not get lost then. Rather these stay fixed and continue to influence/restrict the min-cut). In fact, the algorithm is sensitive to seed point selection. However, due to the interactive nature of the algorithm the user can quickly change the seed point location when not satisfied with the immediate finding and thus optimize the segmentation result.

## Acknowledgements

This work received funding from the European Union in FP7: Clinical Intervention Modelling, Planning and Proof for Ablation Cancer Treatment (ClinicIMPPACT, grant agreement no. 610886) and Generic Open-end Simulation Environment for Minimally Invasive Cancer Treatment (GoSmart, grant agreement no. 600641). Dr. Bernhard Kainz is supported by an EU FP7 MC-IEF 325661 grant and Dr. Xiaojun Chen receives support from NSFC (National Natural Science Foundation of China) grant 81171429. Dr. Dr. Jan Egger receives funding from BioTechMed-Graz ("Hardware accelerated intelligent medical imaging"). The authors would like to thank the clinical staff enabling this study and MeVis in Bremen, Germany, for providing an academic license for the MeVisLab software. Videos demonstrating the interactive segmentation can be found in the following YouTube channel: https://www.youtube.com/c/JanEgger/videos.

## Author Contributions

Conceived and designed the experiments: J.E., H.B. and M.M. Performed the experiments: J.E., H.B. and M.M. Analyzed the data: J.E., H.B. and M.M. Contributed reagents/materials/analysis tools: J.E., H.B., P.B., D.S., M.G., S.S., P.V., M.D., M.H., B.K., A.H., X.C., T.A., M.P., M.M. and D.S. Wrote the paper: J.E., H.B., P.B. and M.P.

## Additional Information

**Competing financial interests:** The authors declare no competing financial interests.

**How to cite this article**: Egger, J. *et al.* Interactive Volumetry of Liver Ablation Zones. *Sci. Rep.* **5**, 15373; doi: 10.1038/srep15373 (2015).